\documentclass[a4paper,conference]{IEEEtran}
\ifCLASSINFOpdf
 \usepackage[pdftex]{graphicx}
 \DeclareGraphicsExtensions{.pdf,.jpeg,.png}
\else
\fi
\usepackage{amsmath}
\usepackage{subfigure}
\usepackage{multirow}
\usepackage[misc]{ifsym}
\usepackage{array}
\usepackage{mathrsfs}
\usepackage{footmisc}

\newcolumntype{C}[1]{>{\centering}p{#1}}
\newcolumntype{L}[1]{>{\raggedleft}p{#1}}
\newcolumntype{R}[1]{>{\raggedright}p{#1}}

\usepackage{url}

\begin{document}
%
\title{FReLU: Flexible Rectified Linear Units for Improving Convolutional Neural Networks}

\author{
	\IEEEauthorblockN{Suo Qiu, Xiangmin Xu and Bolun Cai}
	\IEEEauthorblockA{School of Electronic and Information Engineering,
		South China University of Technology\\
		Wushan RD., Tianhe District, Guangzhou, P.R.China\\
		Email: q.suo@foxmail.com, xmxu@scut.edu.cn, caibolun@gmail.com}
	\and
}

\maketitle

\begin{abstract}
   Rectified linear unit (ReLU) is a widely used activation function for deep convolutional neural networks. However, because of the zero-hard rectification, ReLU networks miss the benefits from negative values.
   In this paper, we propose a novel activation function called \emph{flexible rectified linear unit (FReLU)} to further explore the effects of negative values. By redesigning the rectified point of ReLU as a learnable parameter, FReLU expands the states of the activation output. When the network is successfully trained, FReLU tends to converge to a negative value, which improves the expressiveness and thus the performance. Furthermore, FReLU is designed to be simple and effective without exponential functions to maintain low cost computation.    For being able to easily used in various network architectures, FReLU does not rely on strict assumptions by self-adaption. We evaluate FReLU on three standard image classification datasets, including CIFAR-10, CIFAR-100, and ImageNet. Experimental results show that the proposed method achieves fast convergence and higher performances on both plain and residual networks.

\end{abstract}


%
\IEEEpeerreviewmaketitle

\section{Introduction}

Activation function is an important component in neural networks. It provides the non-linear properties for deep neural networks and controls the information propagation through adjacent layers. Therefore, the design of an activation function matters for the learning behaviors and performances of neural networks. And different activation functions have different characteristics and are used for different tasks. For example, long short-term memory (LSTM) models \cite{lstm1997} use sigmoid or hyperbolic tangent functions, while rectified linear unit (ReLU) \cite{krizhevsky2012imagenet,szegedy2016inception,he2016deep} is more popular in convolutional neural networks (CNNs). In this paper, we mainly focus on extending ReLU function to improve convolutional neural networks.

ReLU \cite{nair2010rectified} is a classical activation function, which effectiveness has been verified in previous works \cite{glorot2011deep,krizhevsky2012imagenet,szegedy2016inception,he2016deep}. The success of ReLU owes to identically propagating all the positive inputs, which alleviates gradient vanishing and allows the supervised training of much deeper neural networks. In addition, ReLU is computational efficient by just outputing zero for negative inputs, and thus widely used in neural networks. 
Although ReLU is fantastic, researchers found that it is not the end of story about the activation function --  the challenges of activation function arise from two main aspects: negative missing and zero-center property.

\textbf{Negative missing.} ReLU simply restrains the negative value to hard-zero, which provides sparsity but results negative missing. The variants of ReLU, including leaky ReLU (LReLU) \cite{maas2013rectifier}, parametric ReLU (PReLU) \cite{he2015delving}, and randomized ReLU (RReLU) \cite{xu2015empirical}, enable non-zero slope to the negative part. It is proven that the negative parts are helpful for network learning. However, non-hard rectification of these activation functions will destroy sparsity.


\textbf{Zero-like property.} In~\cite{clevert2015fast}, the authors explained that pushing the activation means closer to zero (zero-like) can speed up learning. ReLU is apparently non zero-like. LReLU, PReLU, and RReLU cannot ensure a noise-robust negative deactivation state. To this end, exponential linear unit (ELU) \cite{clevert2015fast} was proposed to keep negative values and saturate the negative part to push the activation means closer to zero. Recent variants \cite{trottier2016parametric,li2016improving,carlile2017improving,klambauer2017self,duggal2017p} of ELU and penalized tanh function \cite{xu2016revise} also demonstrate similar performance improvements. However, the incompatibility between ELU and batch normalization (BN) \cite{ioffe2015batch} has not been well treated.



\begin{figure}[tp]
	\centering
	\subfigure[ReLU]{
		\label{fig:relu}
		\includegraphics[width=0.36\linewidth]{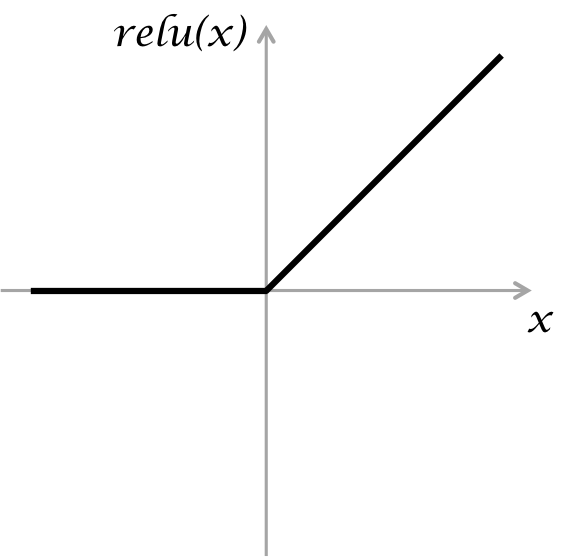}}
	\subfigure[FReLU]{
		\label{fig:frelu}
		\includegraphics[width=0.36\linewidth]{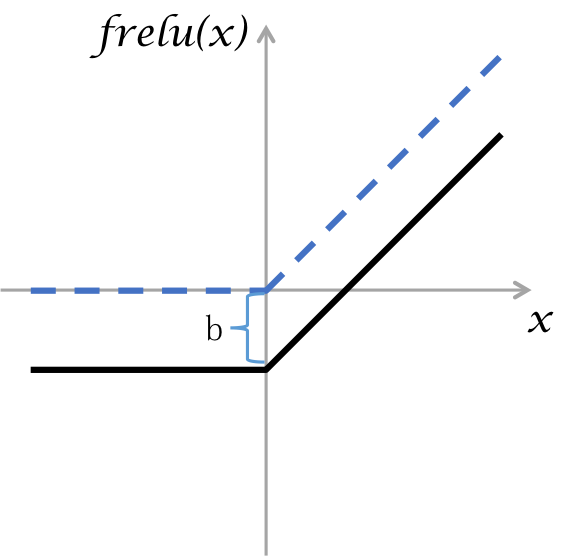}}
	\caption{Illustration of (a) ReLU and (b) FReLU function.}
\end{figure}


In this paper, we propose a novel activation function called \textbf{flexible rectified linear unit (FReLU)}, which can adaptively adjust the ReLU output by a rectified point to capture \textbf{\emph{negative}} information and provide \textbf{\emph{zero-like}} property. We evaluate FReLU on image classification tasks and find that the flexible rectification can improve the capacity of neural networks. In addition, the proposed activation function FReLU brings the following benefits:
\begin{itemize}
\item fast convergence and higher performance;
\item low computation cost without exponential operation;
\item compatibility with batch normalization;
\item weak assumptions and self-adaptation.
\end{itemize}

\section{The Proposed Method}
\subsection{Flexible Rectified Linear Unit}

As illustrated in Fig. \ref{fig:relu}, let variable $x$ represent the input, and rectified linear unit (ReLU) \cite{krizhevsky2012imagenet} is defined as:
\begin{equation} \label{equ:relu}
relu(x) =
\begin{cases}
x & \quad \text{if } x > 0 \\
0 & \quad \text{if } x \le 0
\end{cases}.
\end{equation}

By redesigning the rectified point of ReLU as a learnable parameter, we propose flexible rectified linear unit (FReLU) to improve flexibility on the horizontal and vertical axis, which is expressed as:
\begin{equation} \label{equ:frelu0}
frelu(x) = relu(x + a) + b,
\end{equation}
where $a$ and $b$ are two learnable variables. By further consideration, activation function follows convolutional/linear layer generally, the variable $a$ can be learned together with the bias of the preceding convolutional/linear layer. Therefore, the Equ. \eqref{equ:frelu0} equals to
\begin{equation}
frelu(x) = relu(x) + b,
\end{equation}
which is illustrated in Fig. \ref{fig:frelu}.

Therefore, the \textbf{forward pass} function of FReLU is rewrite as:
\begin{equation}
frelu(x) =
\begin{cases}
x + b_l & \quad \text{if } x > 0 \\
b_l & \quad \text{if } x \le 0
\end{cases},
\end{equation}
where $b_l$ is the $l$-th layer-wise learnable parameter, which controls the output range of FReLU. Note that FReLU naturally generates ReLU when $b_l=0$.

The \textbf{backward pass} function of FReLU is given by:
\begin{equation}
\label{equ:dfrelu}
\begin{array}{l}
\dfrac{\partial frelu(x)}{\partial x} =
\begin{cases}
1 & \quad \text{if } x > 0 \\
0 & \quad \text{if } x \le 0
\end{cases}\\
\dfrac{\partial frelu(x)}{\partial b_l} = 1
\end{array}.
\end{equation}

\subsection{Parameter Initialization with FReLU}

As mentioned in \cite{he2015delving}, it is necessary to adopt appropriate initialization method for a novel activation function to prevent the vanishing problem of gradients. In this subsection, we provide a brief analysis on the initialization for FReLU. More discussions about the initialization of neural networks can refer to \cite{glorot2010understanding,he2015delving}.

\subsubsection{Back propagation}
For the back propagation case, the gradient of a convolution layer is computed by: $\frac{\partial Cost}{\tilde{x}_l} = \hat{W}_l\frac{\partial Cost}{x_l}$, where $x_l = W_l\tilde{x}_l$. $\hat{W}_l$ is a $c$-by-$\hat{n}$ matrix which is reshaped from $W_l$. Here, $c$ is the number of channels for the input and $\hat{n}=k^2d$ ($k$ is the kernel size, and $d$ is the number of channels for the output). We assume $\hat{n}_l$ $w_l$s and $w_l$ and $\frac{\partial Cost}{x_l}$ are independent of each other. When $w_l$ is initialized by a symmetric distribution around zero, $Var\left[\frac{\partial Cost}{\tilde{x}_l}\right] = \hat{n}_lVar[w_l]E\left[(\frac{\partial Cost}{x_l})^2\right]$. And for FReLU, we have: $\frac{\partial Cost}{x_l} = \frac{\partial frelu(x_l)}{\partial x_l} \frac{\partial Cost}{\tilde{x}_{l+1}}$. According to Equ.~\eqref{equ:dfrelu}, we know that $E\left[(\frac{\partial Cost}{x_l})^2\right] = \frac{1}{2}Var[\frac{\partial Cost}{\tilde{x}_{l+1}}]$. Therefore, $Var\left[\frac{\partial Cost}{\tilde{x}_l}\right] = \frac{1}{2}\hat{n}_lVar\left[w_l\right]Var\left[\frac{\partial Cost}{\tilde{x}_{l+1}}\right]$. Then for a network with $L$ layers, we have
$Var\left[\frac{\partial Cost}{\tilde{x}_2}\right] = Var\left[\frac{\partial Cost}{\tilde{x}_{L}}\right]\left(\prod_{l=2}^{L-1}\frac{1}{2}\hat{n}_lVar\left[w_l\right]\right)$.
Therefore, we have the initialization condition:
\begin{equation}
\label{equ:condition}
\frac{1}{2}\hat{n}_lVar\left[w_l\right] = 1\text{, } \forall l,
\end{equation}
which is the same with the msra method \cite{he2015delving} for ReLU.

\subsubsection{Forward propagation}
For the forward propagation case, that is $x_l = W_l\tilde{x}_l$, where $W_l$ is a $d$-by-$n$ matrix and $n=k^2c$. As above, we have $Var[x_l] = n_lVar[w_l]E[\tilde{x}_l^2]$ with the independent assumption. For FReLU, $\tilde{x}_l^2 = \max(0, x_{l-1}^2) + \max(0,2b_lx_{l-1}) + b_l^2$. In general, $x$ is finite or has Gaussian shape around zero, then $E[\tilde{x}_l^2] \approx \frac{1}{2}Var[x_{l-1}]+b_l^2$. Thus, we have
$Var[x_l] \approx (\frac{1}{2}n_lVar[x_{l-1}]+n_lb_l^2)Var[w_l]$.
And for a network with $L$ layers,
$Var[x_L] \approx Var[x_1]\prod_{l=2}^{L}\frac{1}{2}n_lVar[w_l] + \xi$,
where
$\xi = \sum_{k=2}^{L}\left(b_k^2\frac{1}{2^{L-k}}\prod_{l=k}^{L}n_lVar[w_l]\right)$.
We found that the term $\xi$ makes forward propagation more complex. Fortunately when using Equ.~\eqref{equ:condition} for initialization,
$Var[x_L] \approx \frac{c_2}{d_L}Var[x_1] + \sum_{k=2}^{L}\frac{c_k}{d_L}b_k^2$.

In conclusion, when using the initialization condition (Equ. \eqref{equ:condition}) for FReLU, the variance of back propagation is stable and the variance of forward propagation will be scaled by some scalars. FReLU has a relatively stable learning characteristic except in complex applications. Thus, for stable learning, the absolute of $b_l$ prefers to be a small number, especially for very deep models. In practice, by using batch normalization~\cite{ioffe2015batch}, networks will be less sensitive to the initialization method. And the data-driven initialization method LSUV \cite{LSUVInit2015} is also a good choice. For convenience, in this paper, we use MSRA method \cite{he2015delving} (Equ. \eqref{equ:condition}) for all our experiments.

\subsection{Analysis and Discussion for FReLU}
\label{sec:analysis}
In this section, we analyze the improvement of FReLU for neural networks and discuss tips for FReLU.

\subsubsection{State Extension by FReLU}

By adding a learnable bias term, the output range of FReLU $\left[b,+\infty\right)$ is helpful to ensure efficient learning. When $b<0$, FReLU satisfies the principle that activation functions with negative values can be used to reduce bias effect \cite{clevert2015fast}. Besides, negative values can improve the expressiveness of the activation function. There are three output states represented by FReLU with $b<0$:
\begin{equation}
frelu(x) =
\begin{cases}
\text{positive} & \quad \text{if } x > 0 \text{ and } x + b > 0 \\
\text{negative} & \quad \text{if } x > 0 \text{ and } x + b < 0 \\
\text{inactivation} & \quad \text{if } x \le 0
\end{cases}.
\end{equation}


Considering a layer with $n$ units, FReLU with $b=0$ (equal to ReLU) or $b>0$ can only generate $2^n$ output states, while FReLU with $b<0$ can generate $3^n$ output states. Shown in Table \ref{table:init}, the learnable biases tend to negative $b<0$ and bring the improvement in the network by training success. Another factor is that FReLU retains the same non-linear and sparse characteristics as ReLU. In addition, the self-adaptation of FReLU is also helpful to find a specialized activation function.

\subsubsection{Batch Normalization with FReLU}

According to the conclusion in \cite{clevert2015fast} and the experiments in Table \ref{result:smallnet}, PReLU, SReLU, and ELU are not compatible with batch normalization (BN) \cite{ioffe2015batch}. It is because training conflict between the representation restore (scale $\gamma$ and bias $\beta$) in BN and the negative parameter in the activation function. In FReLU, $\max\left(x,0\right)$  isolates two pairs of learnable terms between BN and FReLU. In this paper, we introduce batch normalization (BN) \cite{ioffe2015batch} to stabilize the learning when using the large learning rate for achieving better performance. With BN, backward propagation through a layer is unaffected by the scale of its parameters. Specifically, for a scalar $c$, there is $BN(Wu) = BN((cW)u)$ and thus $\frac{\partial BN((cW)u)}{\partial u} = \frac{\partial BN(Wu)}{\partial u}$. Batch normalization is also a data-driven method, does not rely on strict distribution assumptions. We show the compatibility between BN and FReLU in our experiments (Table \ref{result:smallnet}).

\subsection{Comparisons}
\label{sec:compare}
We compare the proposed FReLU function with a few correlative activation functions, including ReLU, PReLU, ELU, and SReLU.
\begin{figure}[ht]
	\centering
	\includegraphics[width=0.6\linewidth]{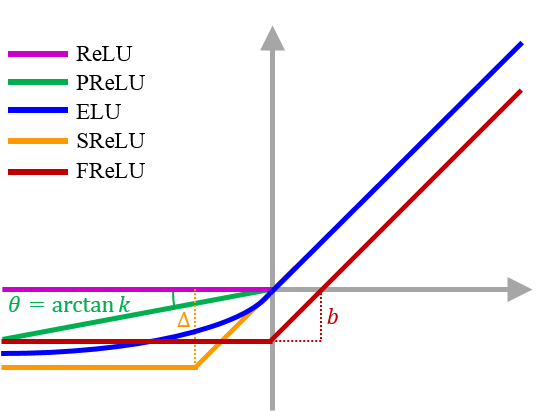}
	\caption{Illustration of the correlative activation functions.}
\end{figure}

\subsubsection{ReLU}
The activation function ReLU \cite{krizhevsky2012imagenet} is defined as $relu(x) = \max(x,0)$. The proposed FReLU function is an extension of ReLU by adding a learnable bias term $b$. Therefore, FReLU retains the same non-linear and sparse properties as ReLU, and extends the output range from $\left[0,+\infty\right)$ to $\left[b,+\infty\right)$. Here, $b$ is learnable parameter for adaptive selection by training. When $b = 0$, FReLU generates ReLU. When $b > 0$, FReLU tends to move the output distribution of ReLU to larger positive areas, which is unnecessary for state extension proven in the experiments. When $b < 0$, FReLU expands the states of the output to increase the expressiveness of the activation function.

\subsubsection{PReLU/LReLU}
The activation function PReLU \cite{he2015delving} is definded as $prelu(x)=\max(x,0) + k * \min(x,0)$, where $k$ is the learnable parameter. When $k$ is a small fixed number, PReLU becomes LReLU \cite{maas2013rectifier}. To avoid zero gradients, PReLU and LReLU propagate the negative input with penalization, thus avoid negative missing. However, PReLU and LReLU probably lose sparsity, which is an important factor to achieve good performance for neural networks. Note that FReLU also can generate negative outputs, but in a different way. FReLU obstructs the negative input as same as ReLU, the backward gradient of FReLU for the negative part is zero and retains sparsity.

\subsubsection{ELU}
The activation function ELU \cite{clevert2015fast} is defined as $elu(x)=\max(x,0)+\min((exp(x)-1),0)$. FReLU and ELU have similar shapes and properties in some extent. Different from ELU, FReLU uses the bias term instead of exponential operation, and reduces the computation complexity. Although FReLU is non-differentiable at $x=0$, the experiments show that FReLU can achieve good performance. In addition, FReLU has a better compatibility with batch normalization than ELU.

\subsubsection{SReLU}
In this paper, shifted ReLU (SReLU) is defined as $srelu(x)=\max(x,\Delta)$, where $\Delta$ is the learnable parameter. Both SReLU and FReLU have flexibility of choosing horizontal shifts from learned biases and both SReLU and FReLU can choose vertical shifts. Specifically, SReLU can be reformed as $srelu(x)=\max(x,\Delta)=\max(x-\Delta,0)+\Delta=\max(x-(\alpha-\Delta)-\Delta,0)+\Delta$, where $(\alpha-\Delta)$ is the learned bias for SReLU. To some extent, SReLU is equivalent to FReLU. In the experiments, we find that SReLU is less compatible with batch normalization and lower performance than FReLU.

\section{Experiments}
\label{sec:exp}

In this section, we evaluate FReLU on three standard image classification datasets, including CIFAR-10, CIFAR-100~\cite{krizhevsky2009learning} and ImageNet \cite{russakovsky2015imagenet}. We conduct all experiments based on \emph{fb.resnet.torch}\footnote{\url{https://github.com/facebook/fb.resnet.torch}} \cite{gross2016training} using the default data augmentation and training settings. The default learning rate is initially set to 0.1. The weight decay is set to 0.0001, and the momentum is set to 0.9. For CIFAR-10 and CIFAR-100, the models are trained by stochastic gradient descent (SGD) with batch size of 128 for 200 epochs (no warming up). The learning rate is decreased by a factor of 10 at 81 and 122 epochs. For ImageNet, the models are trained by SGD with batch size of 256 for 90 epochs. The learning rate is decreased by a factor of 10 every 30 epochs. In addition, the parameter $b$ for FReLU is set to $-1$ as the initialization by default in this paper. For fair comparison and reducing the random influences, all experimental results on CIFAR-10 and CIFAR-100 are reported with the mean and standard deviation of five runs with different random seeds.

\subsection{The Analyses for FReLU}
\label{sec:expressiveness}

\subsubsection{Convergence Rate and Performance}
\begin{figure*}[tp]
	\centering
	\subfigure[Training error]{
		\label{fig:pstr}
		\includegraphics[width=0.238\linewidth]{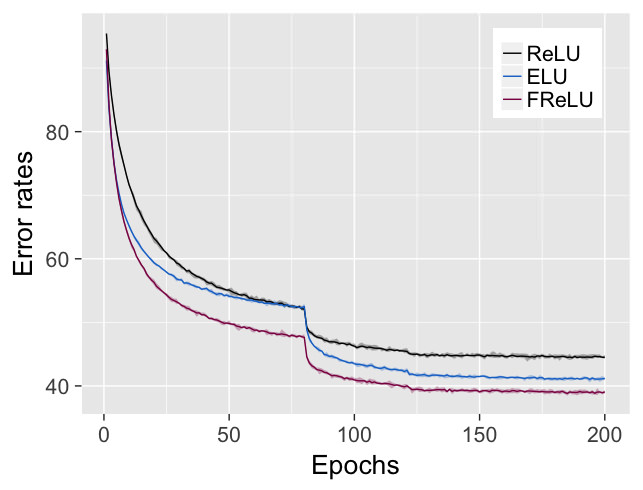}}
	\subfigure[Test error]{
		\label{fig:pste}
		\includegraphics[width=0.238\linewidth]{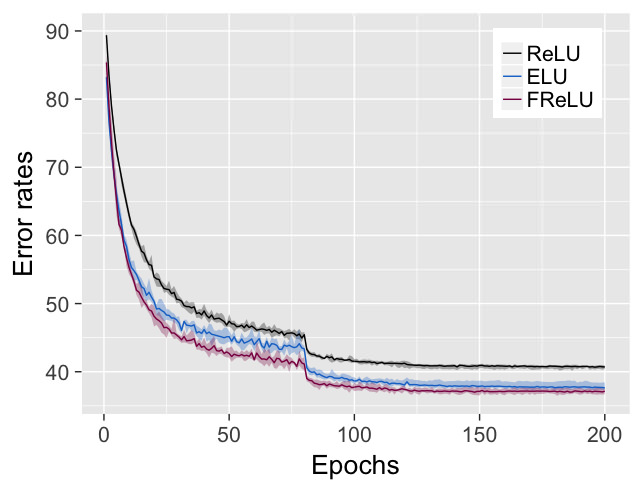}}
	\subfigure[Training error with BN]{
		\label{fig:psbntr}
		\includegraphics[width=0.238\linewidth]{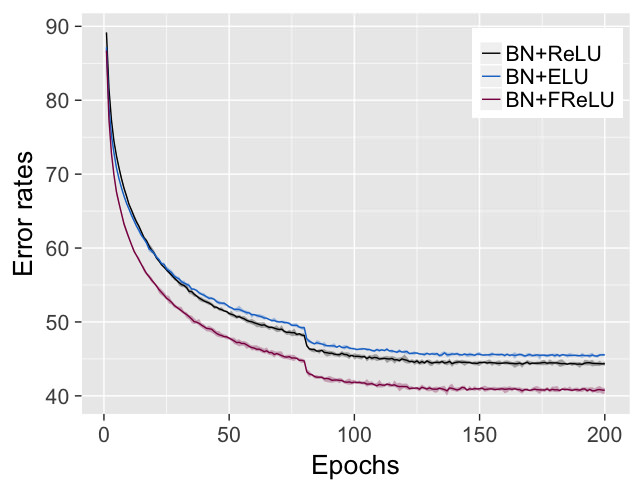}}
	\subfigure[Test error with BN]{
		\label{fig:psbnte}
		\includegraphics[width=0.238\linewidth]{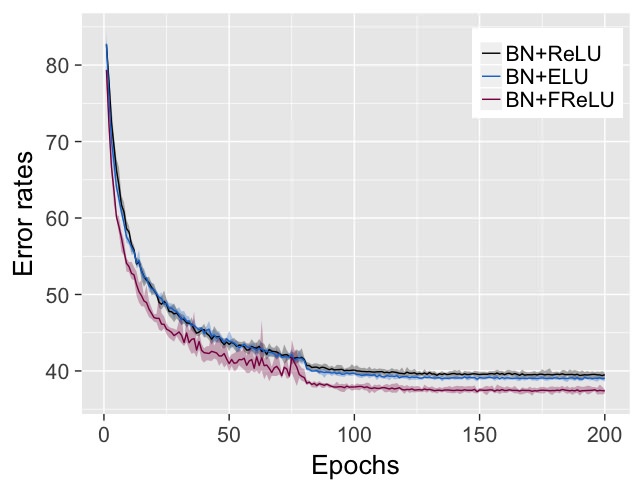}}
	\caption{Error curves on the CIFAR-100 dataset for SmallNet. The base learning rate is 0.01. Best viewed in color.} \label{fig:curves}
\end{figure*}

We firstly evaluate the proposed FReLU on a small convolutional neural network (referred to as SmallNet). It contains 3 convolutional layers followed by two fully connected layers detailed in Table \ref{net:smallnet}. The ACT module is either ReLU, ELU or FReLU. We used SmallNet to perform object classification on the CIFAR-100 dataset \cite{krizhevsky2009learning}. Both training and test error rates are shown in Table \ref{result:smallnet} and we also draw learning curves in Fig. \ref{fig:curves}. We find that FReLU achieves fast convergence and higher generation performance than ReLU, FReLU, ELU, and SReLU. Note that the error rate on test set is lower than training set is a normal phenomenon for a small network on CIFAR-100. 
\begin{table}[tp] 	
	\begin{center}
		\caption{SmallNet architecture on the CIFAR-100 dataset. (BN: Batch Normalization; ACT: activation function.)} \label{net:smallnet}
		\begin{tabular}{|c|c|c|}
			\hline
			\bf{Type} & \bf{Patch Size/Stride} & \bf{\#Kernels} \\
			\hline\hline
			Convolution & 3$\times$3/1 & 32 \\
			\hline
			(BN +) ACT & -- & --\\
			\hline
			MAX Pool & 2$\times$2/2 & --\\
			\hline
			Dropout (20$\%$) & --& --\\
			\hline
			Convolution & 3$\times$3/1 & 64 \\
			\hline
			(BN +) ACT & --& --\\
			\hline
			MAX Pool & 2$\times$2/2 & -- \\
			\hline
			Dropout (20$\%$) & -- & --\\
			\hline
			Convolution & 3$\times$3/1 & 128 \\
			\hline
			(BN +) ACT & --& --\\
			\hline
			MAX Pool & 2$\times$2/2 & --\\
			\hline
			Dropout (20$\%$) &-- &-- \\
			\hline
			Linear & --& 512 \\
			\hline
			(BN +) ACT & --& --\\
			\hline
			Dropout (50$\%$) & --&-- \\
			\hline
			Linear & --& 100 \\
			\hline
			Softmax & --& --\\
			\hline
		\end{tabular}
	\end{center}
\end{table}

\subsubsection{Compatibility with Batch Normalization}
We investigate the compatibilities with batch normalization (BN) on SmallNet. As same in \cite{clevert2015fast}, BN improves ReLU networks but damages ELU networks. We also empirically find that BN does not improve PReLU, SReLU and FReLU when the base learning rate equals to 0.01. No matter with or without BN, FReLU all achieves the lowest testing error rates. Moreover, when using large base learning rate 0.1, ReLU, PReLU, ELU, SReLU, and FReLU networks all cannot converge without BN. With higher learning rates, ReLU, PReLU, and FReLU enjoy the benefits of BN, but ELU and SReLU does not. These phenomenons reflect that FReLU is compatible with BN, which avoids exploding and achieves better performances with large learning rate.
\begin{table*}[tp]
	\begin{center}
		\caption{Comparing ReLU \cite{nair2010rectified}, PReLU \cite{he2015delving}, ELU \cite{clevert2015fast}, SReLU, and FReLU with SmallNet on the CIFAR-100 dataset. We report the mean (std) error results over five runs.}\label{result:smallnet}
		\begin{tabular}{|c|C{2cm}|C{2cm}|C{2cm}|c|}
			\hline
			Base Learning Rate & \multicolumn{2}{c|}{0.01} & \multicolumn{2}{c|}{0.1} \\
			\hline
			Method & Training & Test & Training & ~~ ~ ~~Test~~ ~ ~~ \\
			\hline\hline
			ReLU & 44.20 (0.31) & 40.55 (0.25) & not converge & not converge \\
			PReLU & 42.49 (0.12) & 38.48 (0.33) & exploding & exploding \\
			ELU & 40.79 (0.14) & 37.55 (0.47) & exploding & exploding \\
			SReLU & 39.85 (0.15) & 36.91 (0.17) & exploding & exploding \\
			FReLU & \textbf{38.69 (0.17)} & \textbf{36.87 (0.35)} & exploding & exploding \\
			\hline\hline
			BN+ReLU  & 44.07 (0.18) & 39.20 (0.32) & 42.60 (0.16) & 38.50 (0.43) \\
			BN+PReLU & 42.46 (0.27) & 39.42 (0.54) & 40.85 (0.17) & 37.14 (0.42) \\
			BN+ELU & 45.10 (0.18) & 38.77 (0.18) & 43.27 (0.11) & 37.80 (0.16) \\
			BN+SReLU & 43.47 (0.09) & 38.22 (0.28) & 40.15 (0.07) & 37.20 (0.26) \\
			BN+FReLU & 40.38 (0.26) & 37.13 (0.30) & \textbf{38.83 (0.18)} & \textbf{35.82 (0.12)} \\
			\hline
		\end{tabular}
	\end{center}
\end{table*}

\subsubsection{Different Initialization Values for FReLU}
In this subsection, we further explore the effects of different initialization values for FReLU. We report the results on the CIFAR-100 dataset with the SmallNet. By using a small network, the parameter of FReLU can be fully learned. The test error rates and the convergence values $b$ are shown in Table \ref{table:init}. Interestingly, networks with different initialization values (including positive and negative values) for FReLU are finally converged to close negative value. Assuming the input $x\sim N\left(0,1\right)$, the output expectation of activation function $f\left(x\right)$ can be expressed as $E[x]=\int {\frac{1}{{\sqrt {2\pi } }}\exp \left( { - 0.5{x^2}} \right)f\left( x \right)}$. When the parameter of FReLU $b \approx -0.398$ proven in Table \ref{table:init}, $E[x]$ is approximately equal to zero. Therefore, FReLU is a normalize activation function to ensure the normalization of the entire network.

\begin{table}[tp]
	\begin{center}
		\caption{Mean (std) error results on the CIFAR-100 dataset and convergence values (Layer 1 to 4) for  FReLU with SmallNet.}	\label{table:init}
		\begin{tabular}{|c|c|c|c|c|c|}
			\hline
			Init. Value & Error Rate & Layer1 & Layer2 & Layer3 & Layer4 \\
			\hline \hline
			0.5  & 37.05 (0.07) & -0.3175 & -0.4570 & -0.2824 & -0.3284 \\
			0.2  & 36.71 (0.32) & -0.3112 & -0.4574 & -0.2749 & -0.3314 \\
			0    & 36.91 (0.34) & -0.3144 & -0.4367 & -0.2891 & -0.3313 \\
			-0.4 & 37.10 (0.33) & -0.3235 & -0.4480 & -0.2917 & -0.3315 \\
			-1   & 36.87 (0.35) & -0.3272 & -0.4757 & -0.2849 & -0.3282 \\
			\hline
		\end{tabular}
	\end{center}
\end{table}

\subsubsection{Visualize the Expressiveness of FReLU}
\begin{figure}[tp]
	\centering
	\subfigure[ReLU]{
		\label{fig:embed_relu}
		\includegraphics[width=0.48\linewidth]{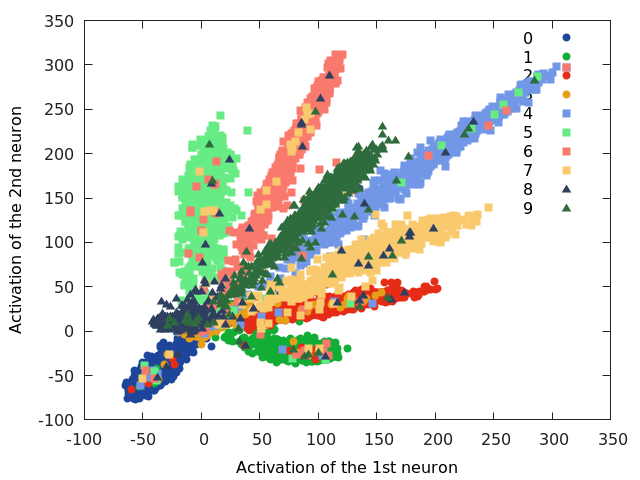}}
	\subfigure[FReLU]{
		\label{fig:embed_frelu}
		\includegraphics[width=0.48\linewidth]{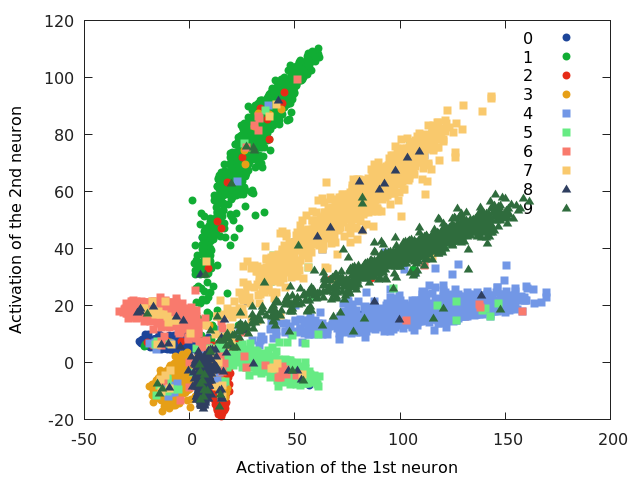}}
	\caption{The distribution of deeply learned features for (a) ReLU and (b) FReLU on the test set of MNIST dataset. The points with different colors denote features from different classes. Best viewed in color.}	\label{fig:embed}
\end{figure}

In order to explore the advantage of FReLU, we further visualize the deep feature embeddings for ReLU and FReLU layers. We conduct this experiment on MNIST \cite{MNIST} dataset with LeNets++\footnote{\url{https://github.com/ydwen/caffe-face/tree/caffe-face/mnist_example}}. As the output number of the last hidden layer in LeNets++ is 2, we can directly plot the features on 2-D surface for visualization. In LeNets++, we use ReLU as the activation function. To visualize the effect of FReLU for feature learning, we only replace the activation function of the last hidden layer as FReLU. We draw the embeddings on the test dataset after training, which are shown in Fig. \ref{fig:embed} and ten classes are shown in different colors. We observe that embeddings of the FReLU network are more discriminative than ReLU's.
The accuracy of the FReLU network is 97.8\%, while the ReLU network is 97.05\%. With negative bias, FReLU provides larger space for feature representation than ReLU.

\subsection{Results on CIFAR-10 and CIFAR-100}

\subsubsection{Results on Network in Network} In this subsection, we compare ReLU, PReLU, ELU, SReLU and FReLU on the Network in Network (referred to as NIN) \cite{lin2013network} model. We evaluate this model on both CIFAR-10 and CIFAR-100 datasets. We use the default base learning rate 0.1 and test with BN. Results are shown in Table \ref{result:nin-cifar}. PReLU seems overfitting and does not obtain good performance. The proposed method FReLU achieves the lowest error rates on the test datasets.
\begin{table}
	\begin{center}
		\caption{Comparing ReLU~\cite{nair2010rectified}, PReLU~\cite{he2015delving}, ELU~\cite{clevert2015fast}, SReLU and FReLU with NIN \cite{lin2013network} model on the CIFAR-10 and CIFAR-100 datasets. The base learning rate is 0.1. We report the mean (std) results over five runs.}\label{result:nin-cifar}
		\begin{tabular}{|c|c|c|c|c|}
			\hline
			Dataset & \multicolumn{2}{c|}{CIFAR-10} & \multicolumn{2}{c|}{CIFAR-100} \\
			\hline
			Method & Training & Test & Training & Test \\
			\hline\hline
			BN+ReLU  & 2.89(0.11) & 8.05(0.15) & 14.11(0.06) & 29.46(0.29) \\
			BN+PReLU & \textbf{1.36(0.03)} & 8.86(0.18) & \textbf{8.96(0.12)} & 33.73(0.29)\\
			BN+ELU & 4.15(0.07) & 8.08(0.26) & 13.36(0.10) & 28.33(0.32) \\
			BN+SReLU & 2.68(0.06) & 7.93(0.24) & 13.48(0.12) & 29.50(0.34) \\
			BN+FReLU & \underline{2.02(0.06)} & \textbf{7.30(0.20)} & \underline{11.40(0.11)} & \textbf{28.47(0.21)}\\
			\hline
		\end{tabular}
	\end{center}
\end{table}

\subsubsection{Evaluation on Residual Networks}
\begin{figure}
	\centering
	\subfigure[Ori. bottleneck \cite{he2016deep}]{
		\label{fig:res-a}
		\includegraphics[width=0.28\linewidth]{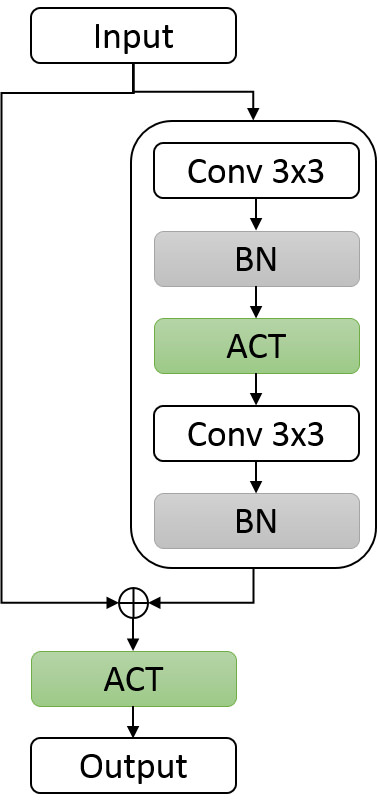}}
	\subfigure[w/o ACT after addition]{
		\label{fig:res-c}
		\includegraphics[width=0.28\linewidth]{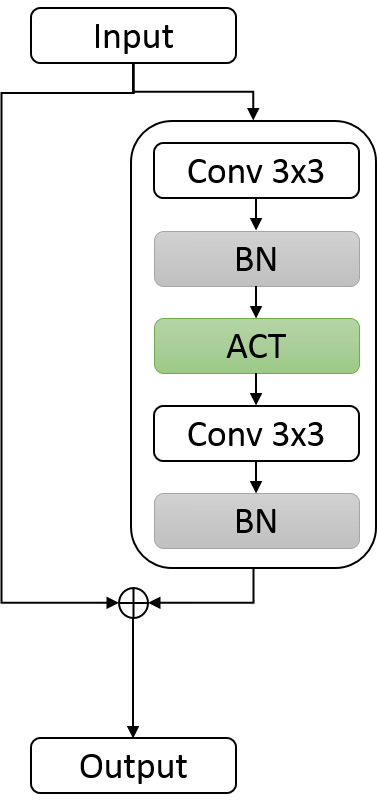}}
	\subfigure[w/o BN after first Conv \cite{elu-resnet}]{
		\label{fig:res-d}
		\includegraphics[width=0.28\linewidth]{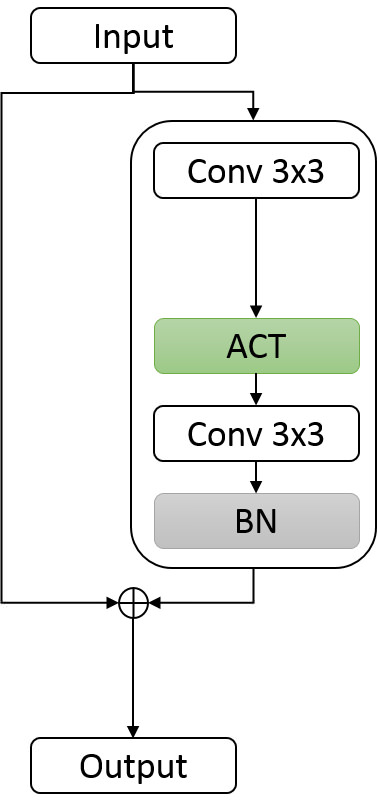}}
	\caption{Various residual blocks.}
\end{figure}

We also investigate the effectiveness of FReLU with residual networks on the CIFAR-10 and CIFAR-100 datasets. Results are shown in Table \ref{table:resnets}. In order to compare the compatibility of FReLU and ELU with BN, we first investigate the performances of residual networks with simply replacing the ReLU with FReLU and ELU, that is using the architecture in Fig. \ref{fig:res-a}. We observe that ELU damages the performances but FReLU improves, which demonstrates that FReLU has the higher compatibility with BN than ELU. Inspired by \cite{elu-resnet}, we further compare the performances with the modified networks, where ELU uses the architecture in Fig. \ref{fig:res-d} and FReLU uses the architecture in Fig. \ref{fig:res-c}. We also observe that FReLU achieves better performances.

\begin{table*}
	\begin{center}
	\caption{Comparing ReLU, ELU ((a)~\cite{clevert2015fast} (c)~\cite{elu-resnet}) and FReLU with ResNet-20/32/44/56/110 \cite{he2016deep} on the CIFAR-10 and CIFAR-100 datasets. We report the mean (std) error rates over five runs.}
	\label{table:resnets}
		\begin{tabular}{|c|c|c|c|c|c|}
			\hline
			\bf{Dataset} & \multicolumn{5}{c|}{\bf{CIFAR-10}} \\
			\hline
			\bf{\#Depths} &
			\multicolumn{1}{c|}{\bf{20}} & \multicolumn{1}{c|}{\bf{32}} & \multicolumn{1}{c|}{\bf{44}} & \multicolumn{1}{c|}{\bf{56}} & \multicolumn{1}{c|}{\bf{110}} \\
			\hline
			Original& 8.12(0.18)  & 7.28(0.19)  & 6.97(0.24) & 6.87(0.54)  & 6.82(0.63) \\
			ELU (a)& 8.04(0.08) & 7.62(0.21) & 7.51(0.22) & 7.71(0.26)  & 8.21(0.21) \\
			FReLU (a)  & 8.10(0.18) & 7.30(0.17)  &  6.91(0.25)  & 6.54(0.22)  & 6.20(0.23) \\
			ELU (c)& 8.28(0.09)  & 7.07(0.17)  & 6.78(0.10)  & 6.54(0.20)  &  5.86(0.14) \\
			FReLU (b) & \textbf{8.00(0.14)} & \textbf{6.99(0.11)}  & \textbf{6.58(0.19)} & \textbf{6.31(0.20)}  & \textbf{5.71(0.19)}\\
			\hline \hline
			\bf{Dataset} & \multicolumn{5}{c|}{\bf{CIFAR-100}} \\
			\hline
			\bf{\#Depths} &
			\multicolumn{1}{c|}{\bf{20}} & \multicolumn{1}{c|}{\bf{32}} & \multicolumn{1}{c|}{\bf{44}} & \multicolumn{1}{c|}{\bf{56}} & \multicolumn{1}{c|}{\bf{110}} \\
			\hline
			Original& 31.93(0.13) & 30.16(0.32)  & 29.30(0.45) & 29.19(0.61) & 28.48(0.85) \\
			ELU (c)& 31.90(0.36) &  30.39(0.37) &  29.34(0.39) & 28.81(0.42) & 27.02(0.32) \\
			FReLU (b)  & \textbf{31.84(0.30)} &  \textbf{29.95(0.27)} & \textbf{29.02(0.25)} & \textbf{28.07(0.47)} & \textbf{26.70(0.38)} \\
			\hline
		\end{tabular}
	\end{center}
\end{table*}

\subsection{Results on ImageNet}
We also evaluate FReLU on the ImageNet dataset. Table \ref{result:nin-imagenet} shows the results with NIN model and a modified CaffeNet, where the result of CaffeNet comes from a benchmark testing \cite{CaffeNetBench2017} and the detailed settings can refer to their project website\footnote{ \url{https://github.com/ducha-aiki/caffenet-benchmark/blob/master/Activations.md}}. FReLU performs well, outperforming other activation functions.
\begin{table}
	\begin{center}
		\caption{Comparing ReLU, ELU and FReLU with NIN model on the ImageNet dataset.}
		\label{result:nin-imagenet}
		\begin{tabular}{|c|c|c|c|}
			\hline
			Network &Method & Top-1 error & Top-5 error \\
			\hline \hline
			\multirow{3}{*}{NIN}&BN+ReLU & 35.65 & 14.53 \\
			&BN+ELU & 38.55 & 16.62 \\
			&BN+FReLU & \textbf{34.82} & \textbf{14.00} \\

            \hline \hline
			\multirow{4}{*}{CaffeNet\footnotemark[3]}&ReLU & 53.00 &--\\
			&PReLU & 52.20 &--\\
			&ELU & \textbf{51.20} &--\\
			&FReLU & \textbf{51.20} &--\\
			\hline
		\end{tabular}
	\end{center}
\end{table}



\section{Conclusion and Future work}
In this paper, a novel activation function called FReLU is proposed to improve convolutional neural networks. As a variant of ReLU, FReLU retains non-linear and sparsity as ReLU and extends the expressiveness. FReLU is a general concept and does not depend on any specific assumption. We show that FReLU achieves higher performances and empirically find that FReLU is more compatible with batch normalization than ELU. Our results suggest that negative values are useful for neural networks. There are still many questions requiring further investigation: (1) How to solve the dead neuron problem well? (2) How to design an efficient activation that can use negative values better and also has better learning property?






\bibliographystyle{IEEEtran}
\bibliography{frelu}

\end{document}